# A Comprehensive Survey on Image Signal Processing Approaches for Low-Illumination Image Enhancement

Muhammad Turab
Norwegian University of Science and Technology, Gjøvik, Norway.

**Abstract**

The usage of digital content (photos and videos) in a variety of applications has increased due to the popularity of multimedia devices. These uses include advertising campaigns, educational resources, and social networking platforms. There is an increasing need for high-quality graphic information as people become more visually focused. However, captured images frequently have poor visibility and a high amount of noise due to the limitations of image-capturing devices and lighting conditions. Improving the visual quality of images taken in low illumination is the aim of low-illumination image enhancement. This problem is addressed by traditional image enhancement techniques, which alter noise, brightness, and contrast. Deep learning-based methods, however, have dominated recently made advances in this area. These methods have effectively reduced noise while preserving important information, showing promising results in the improvement of low-illumination images. An extensive summary of image signal processing methods for enhancing low-illumination images is provided in this paper. Three categories are classified in the review for approaches: hybrid techniques, deep learning-based methods, and traditional approaches. Conventional techniques include denoising, automated white balancing, and noise reduction. Convolutional neural networks (CNNs) are used in deep learning-based techniques to recognize and extract characteristics from low-light images. To get better results, hybrid approaches combine deep learning-based methodologies with more conventional methods. The review also discusses the advantages and limitations of each approach and provides insights into future research directions in this field.

**Index Terms**

Image signal processing, Low-illumination, Image Enhancement, Image Processing, Deep learning

## I. INTRODUCTION

RAW sensor images are transformed into final images by modern cameras using built-in camera processing; these images are usually encoded in a common color space like sRGB. Image adjustments including lens distortion corrections, white balance, noise reduction, sharpening, and color enhancement are made by this processing. Although different camera manufacturers employ different processing methods and stages in their pipelines, these fundamental processes are typically used in most digital camera processing [22]. The primary goal of in-camera processing is to provide images that are both visually appealing and realistically represent the captured scene. However, there are specific challenges with nighttime photography that are not present during the day. For instance, it is usually sufficient to assume the presence of a single illuminant in daylight photography, whereas many illuminants are often present in at night photography. The majority of cameras commonly used for vision systems are incapable of capturing satisfactory nighttime images [21]. Moreover, it is challenging to determine which illuminant should be taken into account during scene color correction due to the unique lighting environment. Mostly artificial light sources can shift suddenly in intensity and direction during nighttime capturing, making it challenging to maintain color balance. Moreover, a night scene might have a wide dynamic range due to the bright light sources like streetlights and the deep shadows in the surrounding region. This makes it difficult to get a well-exposed image that accurately captures the contents of the scene [44].

Deep learning has been widely applied in many domains including image enhancement [23], image generation [26], and image captioning [18]. In image enhancement, it has been very effective for low-light enhancement. Deep learning algorithms for low-light image enhancement offer an enormous advancement in image processing. Traditional approaches like histogram equalization [42], gray level transformations [38], and retinex models [34], often fail to handle the challenges posed by low-light conditions, resulting in noisy, underexposed, and visually unsatisfying images. Deep learning methods, on the other hand, have used neural networks to learn complex mappings between low-light input images and improved equivalents.

This study thoroughly reviews various image signal processing approaches for low-illumination image enhancement. Furthermore, the advantages and disadvantages of each approach are thoroughly discussed. This paper serves as a practical guide for beginners to successfully understand and contribute to this field by presenting the knowledge of image signal processing methods for low-illumination image enhancement. Finally, the open research issues and future directions are discussed. Figure 1 shows the taxonomy of the paper and low-illumination image enhancement approaches. This paper's main contributions are summarized below:

- Review and categorize low-illumination image enhancement approaches into traditional, learning-based, and hybrid methods.

- Discuss the limitations of traditional, learning-based, and hybrid low-illumination image enhancement approaches.
- Overview of various datasets for low-illumination image enhancement and related tasks.
- Identify and discuss open research issues and future directions in low-illumination image enhancement.

The rest of this study is structured as follows. Section II covers the traditional approaches for each category. Section III explores the learning-based approaches in depth. Section IV discusses the hybrid approaches, a combination of traditional and learning-based approaches. Section V discusses the various datasets for low-illumination image enhancement and related tasks. Section VI discusses the open research issues and limitations. Section VII focuses more on the discussion of the objectives, and research problems. Finally, Section VIII concludes the study.

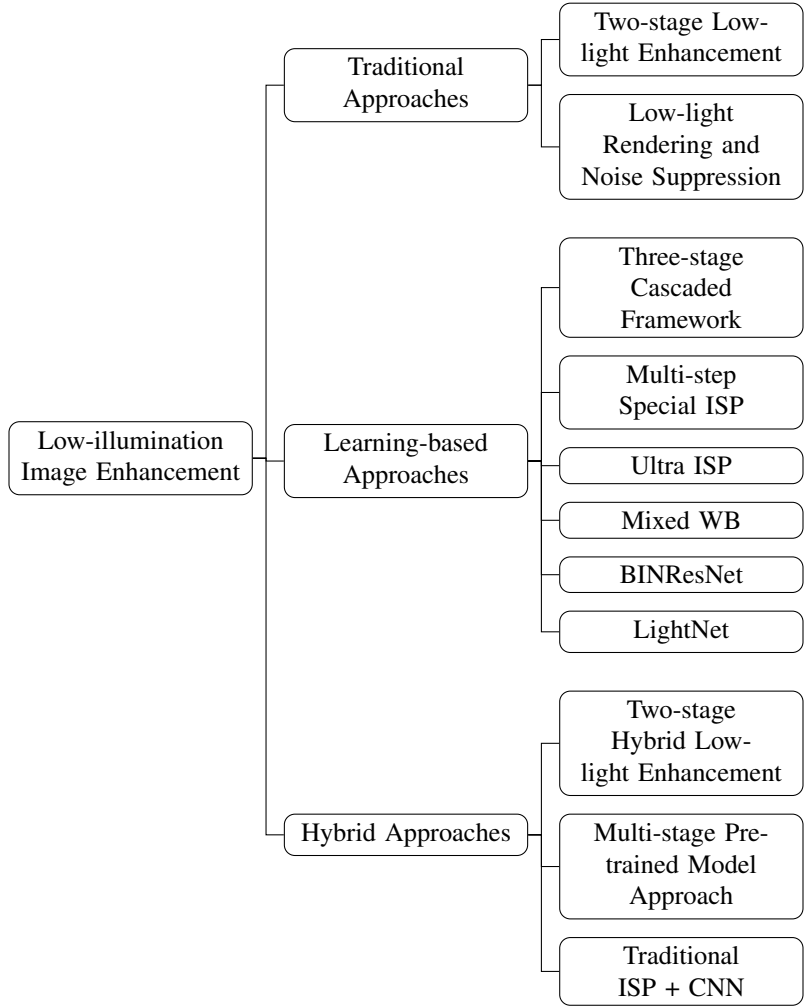

Fig. 1: Taxanomy of the Low-Illumination image enhancement approaches.
**Note:** ISP- Image Signal Processing

## II. TRADITIONAL APPROACHES

An important preprocessing method called low-light image enhancement attempts to enhance an image's visual perception that was taken in low-illumination. Many low-illumination image enhancement methods, from conventional approaches to deep learning-based ones, have been presented over time. The conventional techniques usually involve the simultaneous improvement of brightness, contrast, and noise suppression. The NTIRE challenge featured the application of several conventional methods that produced satisfactory results. Among these techniques are:

### A. Two-stage Low-light Enhancement

The authors propose enhancing low-light images with a two-stage pipeline. The first component works in the RAW domain and consists of demosaicing, automated white balancing via the Gray World method [7], color space conversion to sRGB, and black-and-white level image normalization. This section is consistent with the initial baseline pipeline, except for denoising. The pipeline's second component is designed with low illumination levels in mind. Local Contrast Correction (LCC) is the first step, which improves local contrast without clipping already bright pixels. Saturation and contrast are then improved. Black point correction involves cropping pixels below the 20th percentile value to restore natural aesthetics. After applying a gamma correction with an empirically determined gamma value of 1/1.4, unsharp masking is used to sharpen the image.

After that, the images are expanded to the desired resolution, converted to uint8 format, and then undergo Block-Matching and 3-D Filtering (BM3D) denoising [11]. The denoising intensity is experimentally chosen using noise characteristics found in the image information. To preserve high-frequency information in brighter regions, the denoised picture is blended with the original noisy image using a mask created from the original noisy image's luminance channel. In some situations when the Gray World technique would not work, further automated white balancing is carried out utilizing the Grayness Index [31] to address color casts. After that, the image is rotated using the metadata and saved as a JPEG with a quality of 100.

This method achieves good results, but it has a few limitations. Although this approach has certain limits, it produces good results. First of all, the Gray World Algorithm is straightforward to use with images that meet the Gray World premise. It is less effective, although, with images that defy the gray world assumption, such as those with strong colors or irregular lighting. Additionally, for images with sizable objects of a single hue, the algorithm might produce incorrect results. Furthermore, the BM3D is a powerful and adaptable image-denoising algorithm that produces state-of-the-art results. Its dependency on the image's noise level and processing complexity, however, causes it to fail.

*B. Low-light Rendering and Noise Suppression*

The two main elements of the author's solution—noise reduction and low-light rendering—are based on conventional image processing techniques. They use a multi-stage method based on [40] in the low-illumination rendering phase for the low-light rendering. To provide successful low-illumination improved results, this technique consists of an exposure evaluator, under-exposed recovery, exposure optimization, and exposure fusion. The authors next apply the denoising method known as trilateral weighted sparse coding (TWSC) [39] for noise reduction, which successfully produces denoised results. This two-pronged method creates aesthetically acceptable images in challenging lighting conditions by combining noise reduction with thorough low-light enhancement.

The method achieves promising results, but it has a few limitations. Firstly, the first part uses an algorithm which is a powerful and effective method that achieves state-of-the-art results. It can produce enhanced images with high naturalness and preserve image details, even in challenging lighting conditions. However, it has a few limitations such as its computational complexity, parameter sensitivity, and requirement for multiple input images. In the second part, they use an existing denoising algorithm TWSC which achieves good overall results, especially in the presence of complex noise. It is robust to different types of noise and noise levels and can be applied to a wide range of image types. However, it is more computationally complex than some simpler denoising algorithms.

## III. LEARNING-BASED APPRAOCHES

Deep learning-based approaches, which use a variety of learning algorithms, network architectures, loss functions, training data, etc., have dominated recent developments in low-light image enhancement. Deep learning-based methods have demonstrated excellent results in improving low-light images by successfully reducing noise while keeping important characteristics intact. It has also been demonstrated that these methods work well in handling difficult conditions while improving the quality of augmented images. Promising results were obtained by the majority of the NTIRE challenge methods, which included deep learning algorithms. Here are some of those that are covered:

*A. Three Stage Cascaded Framework*

In their study [33], the authors propose a three-stage cascaded framework to enhance low-light images. This framework sequentially addresses raw image denoising, white balance processing, and Bayer to RGB mapping.

Raw photos are frequently affected by noise in low-light conditions, which degrades their quality. The authors use a denoising network based on the U-Net model [32] to address this problem. This network effectively minimizes noise without losing the image's critical brightness, color, or other characteristics. It is important to note that test and training data have different noise distributions, a problem that is reduced throughout the Bayer to RGB mapping process. The white balance adjustment phase is the next step. The white balance parameters are estimated by the authors using a fully convolutional network and then applied to the image. This approach makes use of training data from the NUS 8-Camera Dataset [9] and the Color Checker Dataset [14]. To accomplish color correction, the image is then multiplied by a predetermined color correction matrix. The authors convert the color-corrected raw image into an RGB image in the last step by using Bayer to RGB mapping. Using a network modified from MW-ISPNet [19] is required for this stage. The network processes the color-corrected raw image as input during testing, resulting in the final RGB image. A high-quality RGB ground truth image is produced by the Bayer to RGB mapping, which also includes other processing stages including denoising, demosaicing, RGB space conversion, tone mapping, and more.

The approach is promising to enhance low-illumination images. The framework sequentially addresses raw image denoising, white balance processing, and Bayer to RGB mapping. The denoising network based on the U-Net model effectively reduces noise without compromising the image's brightness, color, or other essential properties. However, noise distribution differs between training and test data, which can be a challenge during Bayer to RGB mapping. The process involves additional processing steps such as denoising, demosaicing, RGB space conversion, tone mapping, and more, which can be computationally expensive.

## B. Multi-step Special Image Signal Processing

In their study, the authors developed a specialized Image Signal Processing (ISP) pipeline to enhance low-illumination images. This comprehensive pipeline begins with RAW domain denoising using MWRCANet [1], followed by bilinear interpolation for demosaicing. White balance correction is then applied using CAUnet [25], trained on a camera-specific dataset [12]. The resulting image is converted to the CIE-XYZ color space through a color conversion matrix. Tone mapping, crucial for low-illumination image enhancement, is tackled in two ways: self-supervised Unpaired-HDR-TMO [37] and SimuNet trained to simulate human tone mapping preferences. An image evaluation model (ResNet34) selects the optimal tone-mapped output. Lastly, common sense information inherent in low-light images, such as white snow or smoke, is harnessed using Quai-WB [5] to further optimize white balance, yielding the final sRGB output. This ISP pipeline systematically addresses challenges associated with noise reduction, color correction, and tone mapping, resulting in significantly improved low-illumination image quality.

## C. Ultra Image Signal Processing (U-ISP)

The authors introduce the U-ISP approach, aimed at fully leveraging the potential of RAW data. This innovative network design draws inspiration from traditional imaging signal processing (ISP) pipelines, which typically involve varying denoising levels across different channels of the image. To replicate this concept, the authors propose the use of Depth-wise Convolution (DC) Blocks for channel-independent denoising. The DC Block draws inspiration from [29]. Building on recent advancements in image restoration through Channel-wise Transformer [43], the authors incorporate Transposed Self-Attention (TSA) Blocks, inspired [41]. The model architecture aligns with the U-Net framework [32], comprising both an encoder and a decoder. The encoder, primarily responsible for denoising, consists of four stacked DC blocks operating at various scales. Complementing this, the decoder integrates four stacked TSA blocks, addressing restoration tasks across scales corresponding to the encoder settings. In contrast to common practices that involve fixed enhancement factors for low-light image datasets in the RAW domain, the authors adopt a distinct approach to preserve versatility. This is achieved through two key strategies:

- With the MESony100 Dataset, the authors randomly assign enhancement factors ranging from 10 to 100 during training after selecting images with a fixed enhancement factor of 100 from the SID [8] Sony dataset.
- The authors personally took images for the Canon Dataset, which includes a range of scenarios with ISO levels from 100 to 6400. Notably, the Canon and MESony100 Datasets have also undergone rigorous retouching to adhere to accepted standards. The authors then use paired data from their dataset to perform end-to-end training with L1 Loss as the single optimization goal. This novel U-ISP model can produce strong low-light image enhancement in a variety of settings since it is made to fully use RAW data.

## D. Mixed White Balance (Mixed WB)

The authors of this study have proposed a pipeline that seamlessly integrates standard processing stages with advanced techniques for white balance correction and denoising. Specifically, authors incorporate Mixed WB [2], tailored for scenes with mixed lighting conditions, to enhance white balance correction. For denoising sRGB images, authors leverage SwinIR [27], a Swin Transformer [28] based image restoration model. The entire pipeline starts with an input of 16-bit PNG format that has been linearized. Next, the image is normalized using the information that was provided. If any hot pixels are found, they are corrected using interpolation based on nearby pixels in the same color channel. For demosaicking, the Directional Filtering method [30] is utilized rather than the normal CFA interpolation. After that, the pipeline converts raw RGB images to sRGB images colorimetrically. Before converting an image to sRGB, it first transforms it into the CIE XYZ color space. The authors provide Mixed WB, which uses preset white-balance settings and combines calculated weighting maps for correction, to improve white balance correction in mixed-illuminant scenarios. In addition, a memory color enhancement method [6] is incorporated throughout the pipeline to improve skin, sky, grass, and spot colors. After applying gamma correction ($\gamma = 0.8$) to intermediate pictures, the Flash [3] tone mapping operator is used. Image contrast is normalized using an auto-contrast operator that is based on histograms. Noise in night photographs is addressed by the Transformer-based image restoration technique, SwinIR [27], where noise levels are inferred from given information. Due to computational complexity, sub-scaling of images is necessary before denoising. The last stages involve aligning the images according to the orientation of the metadata and then resizing them to the expected challenge output size of 1300 x 866. Finally, to improve edge sharpness in the final sRGB output, unsharp masking is performed on sRGB images.

*E. Batch-Instance Normalization ResNet (BINResNet)*

Ordinance Normalization in Batch The ResNet (BINResNet) model is utilized to improve images. The Cube++ dataset [13] is pre-processed by the authors to get raw, noisy, and white-balanced images, which they then use to train the model. The model's special mix of batch and instance normalization works well to capture the color and structure of images. The standard ResNet [16] generator with batch normalization [20], which is widely utilized in many generation tasks, was unable to capture color information correctly due to the task's intrinsic ill-posedness. While it did a good job of preserving image structure, colors were mostly lost in several areas. Conversely, [36], instance normalization was substituted for batch normalization, preserving colors at the expense of structure. In order to overcome this challenge, the authors divided and concatenated data frequently while utilizing both normalizing procedures in order to preserve color and structure. Initially, the input images are passed through distinct batch and instance normalization residual blocks by the BINResNet model. Each block's outputs are split and concatenated many times to perform the normalization procedure iteratively. Without undergoing any additional normalizing, the final outputs from each normalization layer are sent through a convolution layer to rebuild the 3-channel RGB image. Interestingly, the model predicts white-balanced images with exposure, saturation, and contrast that are randomly modified. After the first training, authors manually adjust these parameters using supplied low-illumination images, allowing the model to produce images efficiently without requiring real night scene data.

*F. LightNet*

The authors of this work provide a generative model that takes into account both local and global information to improve low-illumination images. An encoder-decoder module for gathering global information and a patch discriminator for concentrating on local details—a crucial component in improving image quality—are included in the suggested architecture. The encoder-decoder module facilitates the learning of features at several levels by downsampling the input low-illumination image into many scales. By capturing both local and global feature variations, this method successfully suppresses undesired artifacts like blur and noise. The NTIRE 2022 challenge dataset is used by the authors to assess the effectiveness of their technique, and several quality criteria are used to present the enhancement results. Their method is different from the one described in [24] in that it uses a hierarchical generator in conjunction with a patch discriminator. This ensures that local and global image features are preserved and enhanced, which improves image quality in low-light conditions.

## IV. HYBRID APPROACHES

It has also been investigated if combining traditional and deep learning-based methods could improve the efficiency and usability of low-illumination image enhancement methods. These hybrid approaches usually combine deep learning-based feature extraction and classification algorithms with traditional preprocessing methods. Combining these methods has improved low-illumination images while preserving their quality, with promising results. This section describes these methods.

*A. Two-stage Hybrid Low-light Enhancement*

Low-light scenarios are difficult for traditional image signal processing (ISP) pipelines to adjust to. The authors of this work suggest a two-stage imaging pipeline for low-illumination that combines sophisticated deep-learning-based enhancing techniques with traditional ISP processing. Two steps make up the low-light imaging pipeline: a conventional preprocessing stage and a convolutional neural network (CNN)-based tone augmentation stage. Preprocessing and CNN augmentation are applied sequentially to the input RAW data to create the final improved sRGB image for display. Using modules from the given baseline code, the first stage involves transforming the input RAW image from the linear domain into the sRGB domain. These modules include gamma correction ($\gamma = 2.2$), demosaicing, black level correction (BLC), white level normalization (WLN), demosaicing, white balance (WB) correction, XYZ color space transformation, and white balance (WB) correction. Interestingly, the tone-mapping phase from earlier pipelines is omitted by the authors since it is insufficient to handle the complexity of low-illumination conditions. Rather, they use a large-scale paired night imaging dataset to build an enhancement model. By combining sophisticated deep learning with traditional ISP techniques, this all-encompassing method overcomes the difficulties presented by intricate nighttime situations and produces excellent night photography images.

*B. Multi-stage Pre-trained Model Approach*

The multi-stage method used in the author's solution includes a trained model for tone mapping, white balancing, and denoising. In addition, they apply classic methods like auto-contrast and gamma correction to enhance the brightness and quality of the images, as well as AdaIN (Adaptive Instance Normalization) for color correction. Using the RGGB Bayer pattern, the authors first converted the 1-channel raw data into 4 channels (R, G, B, and G). The pre-trained SID model [14] is then used to handle these 4-channel data to produce RGB output pictures. Every image's mean is computed. The authors use specified thresholds to separate the pictures into two groups to account for changes in the outputs produced by the pre-trained model based on the mean value of the input image. After passing through the pre-trained SID network, images with means lying between 0 and 0.1 are gamma corrected. The goal of this adjustment is to strengthen and elevate the images' visual appeal.

The authors use gamma correction with a coefficient of 0.7 on the raw data for images whose means fall within the range of 0.25 to 0.29. Subsequently, the pre-trained SID network processes the data. AdaIN layer is used to provide color correction by applying color characteristics from a reference image (color cast free) to the denoised image produced by SID. This technique aids in restoring the denoised photographs' inconsistent color. After applying the above methods, the authors rotate the images to the correct orientation and further enhance image quality using auto-contrast and filter sharpening techniques.

*C. Traditional ISP + CNN-based Approach*

The authors' solution pipeline executes picture demosaicing, denoising, white balancing, and tone mapping in order after the basic camera ISP. The initial two procedures are performed using conventional techniques, while the latter two are performed using CNN-based techniques. The RAW image is converted into a three-channel map following the RGGB Bayer Pattern by duplicating the R and B components and averaging the two G components. The researchers used a bilateral filter [35], which is known for its capacity to maintain edges while mitigating noise, to tackle image noise. Based on their radiometric similarity and geographical closeness, this filter computes the average of each pixel. Adjusting the white balance is essential, particularly in environments with poor lighting. The authors use a cascaded structure that shares a backbone and fine-tunes attention at each level, using a state-of-the-art model from [10]. The SimpleCube++ dataset [13] is used to train the model initially, but the authors also use a statistical algorithm [7] to improve the white balance effect for scenes taken at night. The authors employ Adobe Photoshop to colorize white-balanced photos produced from the above stages to create paired data for training the tone mapping network. The original white-balanced photos are used as ground-truth data, and these colorized images are used as inputs for the tone mapping network. To improve the overall visual depiction of the photos, the authors choose to train the tone-mapping model using HDRNet [15].

## V. DATASETS

This section discusses the various datasets for the low-illumination image enhancement tasks.

*A. ColorChecker Dataset*

The ColorChecker dataset comprises 568 high-resolution RAW photos captured by two Canon 1D and 5D camera models. The dataset contains a mix of indoor and outdoor images captured. A MacBeth ColorChecker is included in each image for reference. It is the most extensively used dataset for estimating illumination. Although the ColorCheker dataset is the most widely used dataset for illumination estimation or color constancy, it has a few limitations. On the ColorChecker dataset, illuminant estimation algorithms have been assessed and compared using at least three different ground truths, one of which is different from the other two. Furthermore, all three of these sets of ground truths were estimated inaccurately or incorrectly, in the sense that minor errors were made. In the performance evaluation of illuminant estimate algorithms, the problem of various ground truths and calculation errors has resulted in misleading results. The [17] have found a solution to this issue by introducing a new suggested ground-truth for the ColorChecker dataset.

*B. Nus 8-Camera Dataset*

One of the biggest datasets for color constancy is NUS. It has 1736 raw images. The scenes in this dataset were captured using eight different camera models, with each camera capturing roughly 210 images in total. Typically, tests are run on each camera independently and the mean of all the results is reported, even though the dataset is rather large. Consequently, each experiment uses only 210 images for testing and training, which is insufficient to properly train deep learning-based methods.

*C. Cube+ Dataset*

In [4], the Cube dataset was released. There are 1365 RGB images in this dataset. All of the images in the dataset were taken outdoors in Croatia, Slovenia, and Austria using a Canon EOS 550D camera. Additionally, this dataset was expanded into the Cube+ dataset. There are now 342 additional images in this extension that feature both indoor and outdoor scenes. A calibration item with known surface colors was placed in each scene to calculate the ground-truth illumination. The ground truth distribution of the NUS-8 and the overall distribution of illuminations in the Cube+ are comparable.

*D. INTEL-TAU dataset*

With 7022 high-resolution images, the collection is by far the biggest high-resolution dataset accessible to the general public for training and testing color constancy algorithms. Furthermore, every identifiable face, license plate, and other piece of personal data has been meticulously hidden. The Canon 5DSR, Nikon D810, and mobile Sony IMX135 were the three cameras that were utilized to gather the images included in the INTEL-TAU dataset. The images include sights from farms and laboratories. Real sceneries make up the majority of the dataset, but it also contains a few lab printouts with the corresponding white point data.

## VI. OPEN RESEARCH PROBLEMS

This section focuses on open research challenges in low-illumination image enhancement. Despite recent breakthroughs in low-illumination image enhancement approaches, there are still some unresolved research challenges. Among these issues are:

- **Noise Reduction and Detail retention:**, In low-illumination image enhancement, noise reduction, and detail retention are continuous challenges. While existing approaches have produced promising outcomes, there is still an opportunity for improvement. One potential option is to investigate the use of deep learning-based techniques that may successfully decrease noise while keeping essential characteristics.
- **Robustness to Extreme Conditions:** Another problem in low-illumination image enhancement is robustness under challenging conditions. Current approaches may not perform well in extremely low-light situations or in unusual conditions such as high dynamic range (HDR) images. There is a need for specific solutions that can efficiently improve low-illumination images while keeping their quality in these scenarios.
- **Real-time Processing:** Another challenge that must be addressed in low-illumination image enhancement is real-time processing, particularly in applications such as night vision or surveillance where immediate results are critical. Algorithms that can execute efficiently for real-time tasks need to be developed.
- **Dataset Diversity:** Dataset diversity is another important issue in low-light image enhancement. Diverse and representative datasets are essential for training and evaluating low-light image enhancement models. However, creating such datasets can be challenging due to the lack of ground truth data and the difficulty of capturing images under different lighting conditions. There is a need for larger, more comprehensive collections that can better reflect real-world scenarios.

## VII. DISCUSSION

Low-illumination image enhancement is highly important in various applications both in daily life and industrial use. However, enhancing the images captured under low illumination is full of challenges. Many researchers have tried to use various methods including traditional, and deep learning-based methods to tackle this problem. Traditional approaches are computationally cheaper and provide quick solutions but they are not always accurate due to their limitations as they don't consider the perception of the image as we humans do, rather they just apply a few techniques to reduce the noise, correct the brightness, and saturation. This is exactly what the Two-stage low-light enhancement, low-light rendering, and noise suppression methods focus on. They work well for simple images, where there is no complex structure, and they adjust the noise, brightness, and saturation attributes which might not be suitable. So here comes deep learning, where deep learning models try to learn and mimic the perception of humans. However, that depends on the quality of the data used to train the models. Learning-based models are powerful and can achieve promising results. However, deep learning-based methods continue to have some drawbacks. A significant disadvantage is the over-reliance on a large volume of training data, which can take a long time to collect and annotate. The training process is made more complex by the requirement of appropriately selecting the dataset to ensure that it includes a variety of conditions and classifications. Furthermore, certain deep learning techniques now in use for enhancing low-illumination images largely concentrate on enhancing model performance, sometimes ignoring the importance and generalizability of the results. It is important that these constraints be addressed and that methods that help with more complex tasks be proposed. The goal should be to create methods that can produce excellent results with faster training periods and less training data. Prioritizing the enhanced images' universality and practical use is also important. By addressing these issues, researchers can open the door to more effective and efficient low-light image enhancement techniques that meet the requirements of diverse image processing applications in practical settings.

## VIII. CONCLUSION

In this comprehensive review, various image signal processing techniques for low-illumination image enhancement are discussed. These approaches are classified into three categories: traditional, learning-based, and hybrid. Related datasets, open research issues, and future directions are also discussed. The analysis shows that traditional methods are simple and computationally efficient but lack the ability to handle complex image structures. Learning-based methods, on the other hand, are more effective in handling complex structures but require large amounts of training data. Hybrid methods combine the strengths of both traditional and learning-based methods and have shown promising results. The key takeaway can be summarized as deep learning-based methods hold the potential to enhance low-illumination images in real-time applications with the only concern of making the models faster and more efficient.